\documentclass[runningheads]{llncs}

\usepackage{eccv}

\usepackage{eccvabbrv}

\usepackage{graphicx}
\usepackage{booktabs}
\usepackage{multirow}
\usepackage{placeins}

\usepackage[accsupp]{axessibility}  

\usepackage{hyperref}

\usepackage{orcidlink}
\newcommand{\methodname}{GO-Renderer\xspace}

\begin{document}

\title{GO-Renderer: Generative Object Rendering with 3D-aware Controllable Video Diffusion Models}

\titlerunning{GO-Renderer}

\author{
    Zekai Gu\inst{1,2} \and
    Shuoxuan Feng\inst{3} \and
    Yansong Wang\inst{4} \and
    Hanzhuo Huang\inst{5} \and
    Zhongshuo Du \and
    Chengfeng Zhao\inst{1} \and
    Chengwei Ren\inst{1} \and
    Peng Wang\inst{2}\thanks{Project leader.} \and
    Yuan Liu\inst{1}\thanks{Corresponding author.}
}

\authorrunning{Z. Gu et al.}

\institute{
    $^1$HKUST \quad
    $^2$VAST \quad
    $^3$Nanyang Technological University \\
    $^4$Tsinghua University \quad
    $^5$ShanghaiTech University
}

\maketitle

\begin{abstract} 
Reconstructing a renderable 3D model from images is a useful but challenging task.
Recent feedforward 3D reconstruction methods have demonstrated remarkable success in efficiently recovering geometry, but still cannot accurately model the complex appearances of these 3D reconstructed models. 
Recent diffusion-based generative models can synthesize realistic images or videos of an object using reference images without explicitly modeling its appearance, which provides a promising direction for object rendering, but lacks accurate control over the viewpoints.
In this paper, we propose \methodname, a unified framework integrating the reconstructed 3D proxies to guide the video generative models to achieve high-quality object rendering on arbitrary viewpoints under arbitrary lighting conditions. 
Our method not only enjoys the accurate viewpoint control using the reconstructed 3D proxy but also enables high-quality rendering in different lighting environments using diffusion generative models without explicitly modeling complex materials and lighting.
Extensive experiments demonstrate that \methodname achieves state-of-the-art performance across the object rendering tasks, including synthesizing images on new viewpoints, rendering the objects in a novel lighting environment, and inserting an object into an existing video. Project page: \href{https://igl-hkust.github.io/GO-Renderer}{https://igl-hkust.github.io/GO-Renderer}

\vspace{-5mm}
\end{abstract}

\begin{figure}
    \centering
    \includegraphics[width=0.999\linewidth]{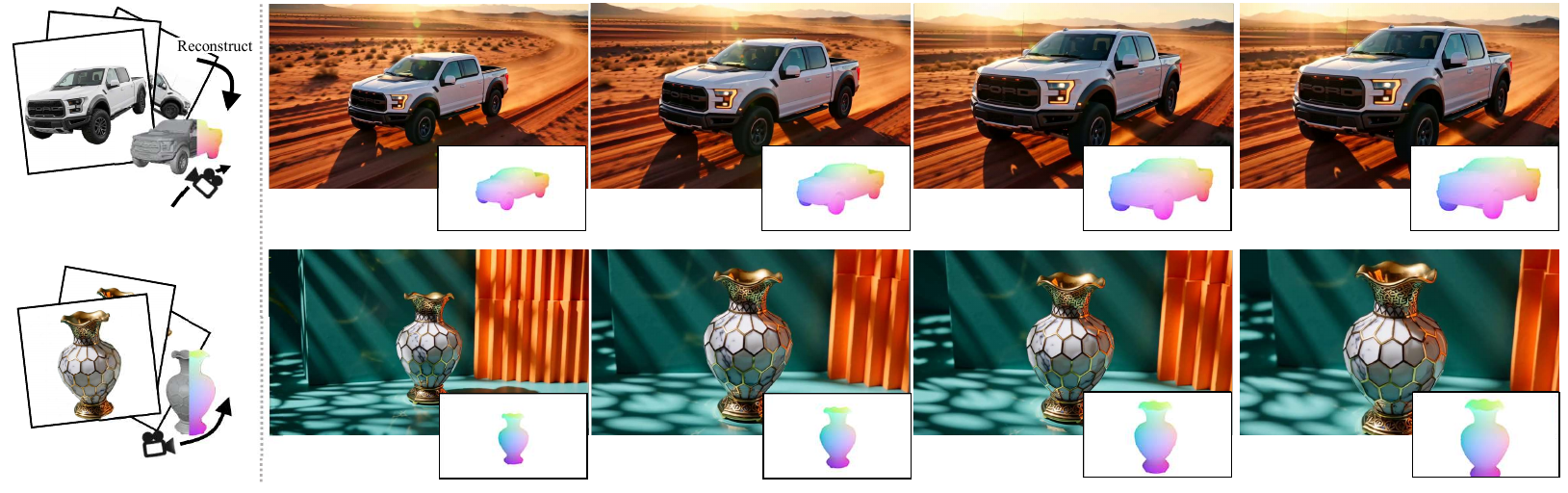}
    \caption{Results of our \methodname. Driven by multi-view references and 3D poses, \methodname renders multi-view consistent object videos seamlessly integrated into novel environments along arbitrary camera trajectories.}
    \vspace{-7mm}
    \label{fig:teaser}
\end{figure}

\section{Introduction}
\label{sec:intro}

Rendering high-fidelity objects within diverse environments is a fundamental pursuit with extensive applications in advertising, film production, and immersive content creation. Achieving realistic object relighting and novel view synthesis requires not only reconstructing precise 3D geometry from 2D observations but also accurately simulating complex light-transport phenomena. Fundamentally, this relies on inverse rendering, the process of disentangling intrinsic scene properties (i.e., geometry, illumination, and materials) from 2D images. However, because shape, texture, and lighting are inextricably coupled in the observed radiance, decoupling illumination from surface materials constitutes a highly ill-posed problem. Consequently, acquiring accurate and multi-view consistent Physically Based Rendering (PBR) materials from standard imagery remains a formidable challenge, often limiting the physical correctness of rendered objects under drastic viewpoint and environmental changes.

To acquire a renderable object model suitable for diverse environments, mainstream solutions generally follow a ``reconstruct-then-render'' paradigm. These approaches can be broadly categorized into traditional mesh-based pipelines and recent radiance-based neural rendering methods. Traditional multi-view reconstruction techniques~\cite{schoenberger2016mvs,schoenberger2016sfm,pan2024glomap} and contemporary generative 3D models~\cite{hunyuan3d,trellis,reconviagen,triposg} are highly effective at establishing robust geometric scaffolds. However, relying on these geometries to further recover complex Physically-based Rendering (PBR) materials involves brittle, multi-stage optimizations, making it exceedingly difficult to restore accurate, high-fidelity surface materials. Conversely, neural rendering paradigms like NeRF~\cite{nerf} and 3D Gaussian Splatting (3DGS)~\cite{3dgs} excel at Novel View Synthesis (NVS) for specific captured scenes. Yet, they inherently bake the original environmental illumination into their representations. Consequently, rendering these objects in novel lighting environments is non-trivial and necessitates complex inverse rendering to disentangle the baked lighting. In essence, while recovering coarse 3D geometry has become increasingly tractable, especially with the emergence of feedforward 3D reconstruction models~\cite{anysplat,volsplat,vggt,pi3,gen3r,flashworld,mvsplat}, accurately modeling and decoupling surface appearance remains a critical bottleneck across both traditional and neural reconstruction frameworks, severely hindering the acquisition of fully renderable object models.

A highly promising direction has emerged with reference-based video diffusion models~\cite{phantom, magref, storymem, skyreels}. By leveraging the robust visual priors encapsulated within foundation models, these approaches can directly synthesize high-fidelity videos of a target object under diverse, complex lighting conditions based simply on a 2D reference image. This elegantly bypasses the notoriously difficult process of explicitly decoupling physical materials and illumination. However, despite their impressive generative capabilities, these models are still far from serving as reliable object renderer. First, they typically rely on a single-view reference; consequently, the appearance of unseen regions is left entirely to the model's unconstrained hallucination, often destroying multi-view consistency. Second, they inherently lack precise viewpoint controllability, making it exceedingly difficult to dictate strict camera trajectories. In summary, while reference-based diffusion models successfully avoid explicit physical modeling and achieve high-quality video synthesis, their inherent lack of spatial determinism and multi-view constraints prevents them from functioning as a controllable object renderer.

In this paper, by combining explicit 3D representations with expressive video generation models, we introduce GO-Renderer (Generative Object Renderer), a novel framework designed to construct a fully controllable object renderer from sparse images. Given a few reference images of a target object, a desired relative camera trajectory, and a novel environmental illumination (specified via texts, images, or videos), GO-Renderer synthesizes high-fidelity, multi-view consistent videos for the object. 

The core insight of our approach lies in leveraging a fast, albeit coarse, geometric reconstruction, like ReconViaGen~\cite{reconviagen} or VGGT~\cite{vggt}, to build a structural 3D proxy. This 3D proxy not only acts as a robust geometric guidance signal to enforce precise, spatially deterministic viewpoint control within a video diffusion model. Meanwhile, it also functions as a spatial bridge to seamlessly aggregate features from multiple reference images, effectively mitigating the hallucination issues of single-view generation and ensuring strict multi-view consistency. By marrying the explicit spatial constraints of a 3D proxy with the powerful visual priors of generative models, GO-Renderer elegantly bypasses the notoriously ill-posed physical modeling of complex materials and lighting. Our method enables highly controllable, photorealistic novel view and lighting synthesis, establishing a highly promising new direction for generative object rendering.

To explicitly realize this conceptual 3D proxy within a generative framework, our primary technical contribution lies in formulating the geometric guidance of 3D proxy as object-centric coordinate maps. Specifically, we render the coarse 3D reconstruction into dense coordinate maps for both the input reference views and the target camera trajectory. In these coordinate maps, the color of each pixel directly encodes its 3D coordinates in the object's coordinate system. We then channel-wise concatenate the target coordinate map with the latent noise, while similarly pairing the reference images with their corresponding reference coordinate maps. By feeding these spatially aligned conditions into the diffusion model, we establish a dense, pixel-level correspondence. The diffusion model essentially learns to ``look up'' the correct appearance from the reference images based on shared 3D coordinates. This elegant design not only enforces strict spatial constraints for precise viewpoint control but also guarantees highly consistent texture transfer across varying camera poses.

Finally, training a robust object renderer requires large-scale data containing multi-view reference images paired with target videos under varying environmental lighting and known camera poses. Existing datasets (e.g., CO3D~\cite{co3d}, OmniObject3D~\cite{omniobject3d}) typically feature static illumination and lack the necessary lighting variations. To overcome this critical data scarcity, we constructed a comprehensive large-scale dataset. By leveraging advanced video generation models and synthetic rendering pipelines, we synthesized 57k high-quality video clips with diverse lighting conditions and corresponding proxy geometries, providing a robust foundation for training our generative renderer.

We extensively evaluate GO-Renderer across two challenging settings: object video synthesis under novel environmental illumination and novel view synthesis (NVS) under original lighting. Our method significantly outperforms current state-of-the-art diffusion~\cite{videoxfun} or reconstruction methods~\cite{anysplat,reconviagen}. Furthermore, we demonstrate the compelling practical utility of GO-Renderer through downstream applications like seamless video object insertion, where our method naturally harmonizes the target object with complex, dynamic backgrounds while maintaining strict geometric and illumination consistency.

\section{Related Work}
\label{sec:related}

\subsection{Reconstruction and Rendering}
While recent image-domain methods have explored 3D integration—such as RefAny3D~\cite{refany3d}, which uses dual-branch perception to condition image generation on 3D asset point maps, AGP~\cite{sweetdreamer}, which fine-tunes 2D generation via viewpoint-specific coordinate maps, thereby maintaining the pure generative nature of appearance synthesis without explicit 3D optimization, and IC-Light~\cite{iclight}, which parses lighting priors to relight 3D scenes via Gaussian Splatting—mainstream object-centric video rendering still heavily relies on explicitly reconstructing the 3D model of the object first.

\textbf{3D Object Reconstruction.} 3D reconstruction methods have long dominated novel-view synthesis. NeRF~\cite{nerf} pioneered the use of continuous volumetric scene functions optimized via MLPs for view synthesis, while 3D Gaussian Splatting (3DGS)~\cite{3dgs} revolutionized the field by enabling real-time rendering through interleaved optimization of 3D Gaussians with anisotropic covariance. To eliminate the need for costly per-scene optimization, feed-forward networks have been proposed. Models like AnySplat~\cite{anysplat} and VOLSplat~\cite{volsplat} predict Gaussian primitives directly from uncalibrated image collections in a single forward pass. Similarly, VGGT~\cite{vggt} and DepthAnything3~\cite{depthanything3} leverage transformer backbones to infer 3D attributes (camera parameters, point maps) from arbitrary visual inputs, and Pi3~\cite{pi3} introduces a permutation-equivariant architecture to predict affine-invariant poses without requiring a fixed reference view. On the generative side~\cite{hunyuan3d,triposg,trellis} synthesize high-fidelity 3D meshes and structured latent representations from images or text. Additionally, ReconViaGen~\cite{reconviagen} integrates multi-view reconstruction priors with diffusion-based generation to address the hallucination constraints and inconsistencies of pure generative models.

Despite these advancements, a critical limitation remains: while explicit methods guarantee multi-view geometric consistency, the reconstruction process inevitably causes a loss of precision, texture details, and material properties. For instance, explicit representations like 3DGS~\cite{3dgs} struggle with dynamic relighting and interactive editing, often leading to degradation of high-frequency texture details. Conversely, purely generative reconstruction methods~\cite{reconviagen} often produce outputs with inevitable appearance and geometric inconsistencies.

\textbf{Rendering and Relighting.} Once a 3D model is obtained, generative methods are widely employed to render the subject in novel scenes with distinct strategies: DiffusionRenderer~\cite{diffusionrenderer} estimates G-buffers for photorealistic forward rendering; Diffusion as Shader~\cite{das} leverages 3D tracking videos to impose spatial constraints; UniLumos~\cite{unilumos} and Light-A-Video~\cite{lightavideo} focus on geometry-guided correction and training-free temporally smooth relighting, respectively; while RelightMaster~\cite{relightmaster} and RelightVid~\cite{relightvid} utilize Multi-plane Light Images and in-the-wild augmentations to tackle paired-data scarcity. Ultimately, the performance of these cascaded approaches remains strictly bottlenecked by the precision of the preceding 3D reconstruction stage.

\subsection{Video Diffusion Models}
The rapid evolution of video diffusion models encompasses both foundational base models and methods tailored for controllable video generation.
ecent breakthroughs in video foundation models have significantly pushed the boundaries of generation through the widespread adoption of Diffusion Transformer (DiT) architectures. Various models have emerged with distinct and complementary capabilities: Wan~\cite{wan} demonstrates the scaling laws of video generation with up to 14 billion parameters for state-of-the-art visual performance; LongCat~\cite{longcat} and Open-Sora-Plan~\cite{opensoraplan} excel in efficient, high-resolution long video generation; Hunyuan Video~\cite{hunyuanvideo} optimizes consumer-grade efficiency while maintaining highly coherent motion; and LTX-Video~\cite{ltx2} extends the generative scope by introducing native synchronized audiovisual generation.

\textbf{Geometry-aware Controllable Generation.} Building upon these foundation models, recent frameworks have increasingly incorporated 3D geometric conditions to achieve precise spatial and camera control. For instance, GEN3C~\cite{gen3c} integrates 3D-informed representations to maintain world consistency under explicit camera trajectories. ViewCrafter~\cite{viewcrafter} leverages point cloud renderings to guide video diffusion models, facilitating high-fidelity Novel View Synthesis (NVS). Similarly, Diffusion as Shader (DaS)~\cite{das} formulates the diffusion process as a neural shader, utilizing 3D-aware conditions (e.g., geometry and lighting) to control versatile video generation. 

While these approaches demonstrate capabilities in scene-level novel view synthesis, they are sub-optimal for object-centric generative rendering. These methods predominantly condition on single images or temporally adjacent frames, limiting their capacity to extract and fuse comprehensive appearance priors from unconstrained multi-view references. Consequently, when synthesizing a specific object under complex, arbitrary camera trajectories within entirely novel environments, they struggle to maintain strict multi-view consistency and preserve fine-grained textures without resorting to explicit, lossy appearance modeling.

\textbf{Subject-driven Generation.} Subject-driven generation seeks to end-to-end synthesize videos of a specific subject in novel environments, typically conditioned on a single object view. Recent advances highlight various unique capabilities: Phantom~\cite{phantom} aligns text-image-video triplets to prevent multi-subject confusion; MAGREF~\cite{magref} utilizes region-aware masked guidance to alleviate copy-paste artifacts; and StoryMem~\cite{storymem} maintains cross-shot storytelling consistency via a visual memory bank and negative RoPE shifts. However, relying purely on single-view references inherently restricts their ability to precisely control viewpoint transformations. Due to the limited 3D geometric comprehension of foundational 2D models, generating dynamic viewpoint changes in these methods frequently results in severe geometric distortions and anatomically incorrect content.

\section{Methodology}
\label{sec:methodology}

\subsection{Overview}
\label{sec:overview}

\begin{figure}[t!]
    \centering
    \includegraphics[width=0.99\linewidth]{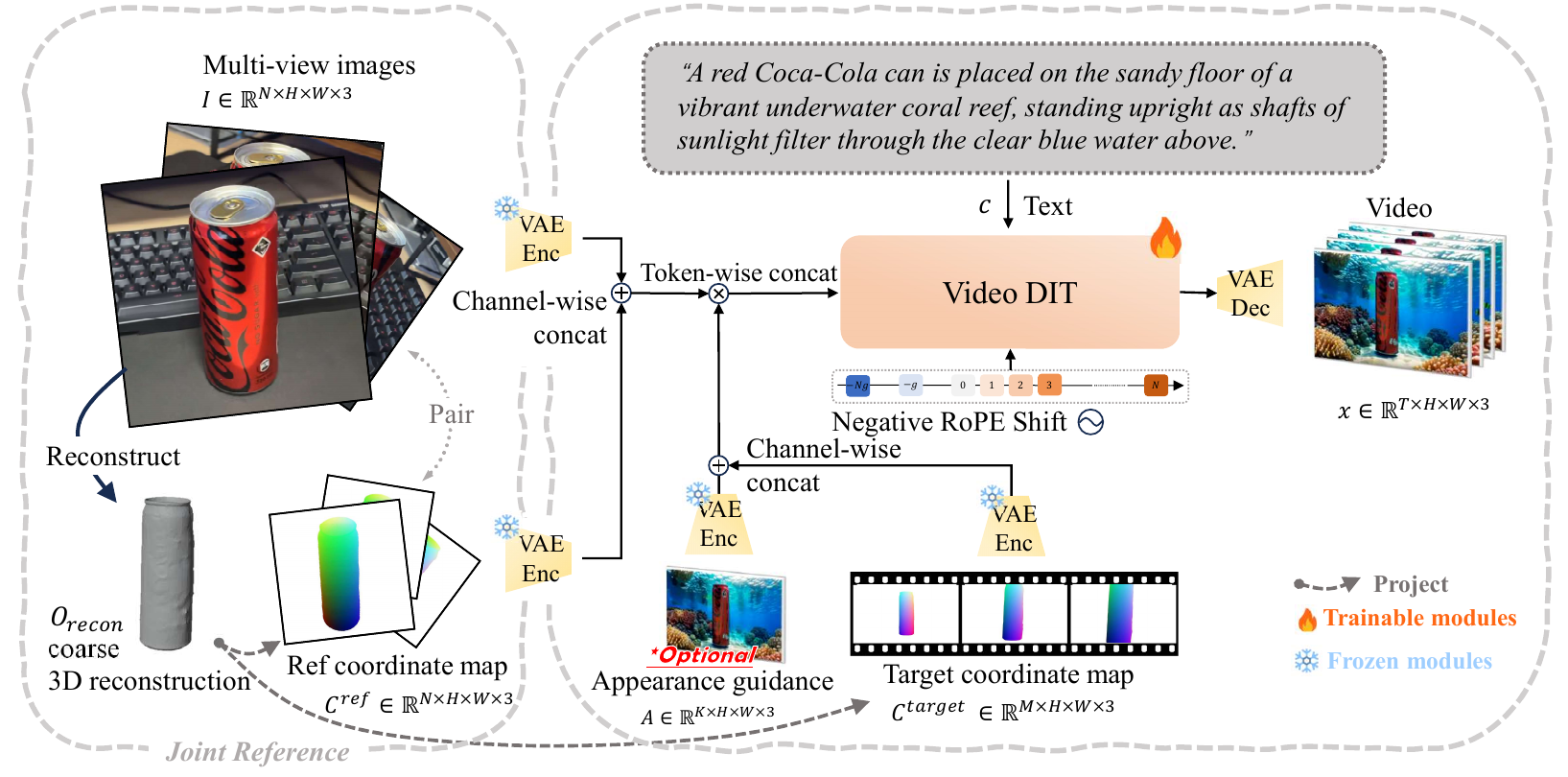}
    \caption{\textbf{Overview of \methodname.} Given multi-view reference images, we first establish a coarse 3D geometry via 3D reconstruction to render reference and target coordinate maps. These maps bridge explicit 3D geometry with the multi-view reference images via video diffusion model. Specifically, the reference images and their coordinate maps are paired via channel-wise concatenation, while the target coordinate maps are combined with optional appearance guidance. These spatial conditions are then jointly processed by a text-conditioned Video Diffusion Transformer to render high-fidelity object videos with precise viewpoint control.}
    \label{fig:pipeline}
      \vspace{-3mm}
\end{figure}

We aim to construct an object renderer for rendering high-fidelity object videos with precise viewpoint control from sparse visual inputs. Explicit 3D reconstruction pipelines can establish robust geometry but struggle with the extraction of complex surface materials, whereas reference-based video diffusion models can synthesize realistic illumination but inherently lack multi-view consistency. To overcome these limitations, we propose a hybrid framework that synthesize the explicit spatial constraints of a coarse 3D proxy with the powerful visual priors of generative models.

Specifically, given a set of reference images $\mathcal{I} = \{I_j\}_{j=1}^N$ of an object, a sequence of target poses $\mathcal{P}^\text{target} = \{P_k^\text{target}\}_{k=1}^M$, text prompts $c$ and optional appearance guidance videos $\mathcal{A}$ which describe the lighting environment, our target is to render a video $x$ of the object corresponding to the given target poses in the given lighting environment
\begin{equation}
    x=\mathcal{R}_{\text{GO-Renderer}}(\mathcal{I}, \mathcal{P}^\text{target}, \mathcal{A}, c).
\end{equation}

To achieve this, we first run a fast 3D reconstruction method, e.g. ReconViaGen~\cite{reconviagen} or VGGT~\cite{vggt}, to get the camera poses $\mathcal{P}$ for all reference images and a coarse 3D reconstruction $\mathcal{O}_{\text{recon}}$ which is either a mesh or a 3D point cloud. Then, we will construct a 3D proxy from the 3D reconstruction (Sec.~\ref{sec:pose_def}) and utilize the constructed 3D proxy as condition to guide a video diffusion model to generate the corresponding object videos (Sec.~\ref{sec:architecture}). 

\subsection{Pose Representation via 3D Proxy}
\label{sec:pose_def}

In this section, we construct a 3D proxy that serves as the condition for the subsequent video diffusion model. This constructed 3D proxy fulfills a critical dual purpose. First, it acts as a robust geometric guidance signal to enforce precise, spatially deterministic viewpoint control over the generated video sequence. Second, it functions as a spatial bridge to seamlessly aggregate appearance features from the multi-view reference images $\mathcal{I}$, effectively mitigating the hallucination issues inherent in single-view generation and ensuring strict multi-view consistency across the rendered frames.

Specifically, we formulate this 3D proxy using object-centric coordinate maps. Given the coarse 3D reconstruction $\mathcal{O}_{\text{recon}}$ and the estimated reference camera poses $\mathcal{P} = \{P_j\}_{j=1}^N$, we rasterize $\mathcal{O}_{\text{recon}}$ to render a set of reference coordinate maps $\mathcal{C}^{\text{ref}} = \{C_j^{\text{ref}}\}_{j=1}^N$. Similarly, we project $\mathcal{O}_{\text{recon}}$ onto the desired target camera trajectory $\mathcal{P}^{\text{target}} = \{P_k^{\text{target}}\}_{k=1}^M$ to obtain the corresponding target coordinate maps $\mathcal{C}^{\text{target}} = \{C_k^{\text{target}}\}_{k=1}^M$. In these coordinate maps, the RGB color value of each valid foreground pixel directly encodes its normalized $(x, y, z)$ spatial coordinates within the object's local coordinate system.

\subsection{Architecture}
\label{sec:architecture}

In this section, we introduce a Joint Reference Structure to integrate the aforementioned conditional variables into the video diffusion model. By leveraging the correspondence between the reference images $\mathcal{I}$ and the coordinate maps $\mathcal{C}^{\text{ref}}$, this mechanism accurately transfers the appearance features of the references $\mathcal{I}$ to the rendered video $x$ across different target coordinate maps $\mathcal{C}^{\text{target}}$. Additionally, we introduce a Negative RoPE Shift mechanism to temporally isolate the spatial references from the target video sequence $x$, preventing transition artifacts.

\textbf{Joint Reference Structure.}
As shown in the right part of Fig.~\ref{fig:pipeline}, to construct the generation conditions using these coordinate maps, we pair the multi-view reference images $\mathcal{I}$ with their corresponding reference coordinate maps $\mathcal{C}^{\text{ref}}$ along the channel dimension. Concurrently, during the iterative denoising process, the target coordinate maps $\mathcal{C}^{\text{target}}$ and the optional videos $\mathcal{A}$ describing the lighting environment are concatenated with the target latent noise along the channel dimension. By feeding these spatially aligned conditions into the diffusion model, we establish a dense, pixel-level correspondence between the source references and the target generation. Because the object's local coordinates are invariant to camera transformations, the diffusion model learns to "\textit{look up}" the appearance and texture details from the reference views $\mathcal{I}$ based on the shared 3D coordinates encoded in $\mathcal{C}^{\text{ref}}$ and $\mathcal{C}^{\text{target}}$. This mechanism ensures consistent texture transfer across varying camera poses without requiring explicit physical modeling of surface materials.

\textbf{Negative RoPE Shift.} 
Video diffusion models normally apply temporal positional embeddings (PE) to the input sequence. If the reference sequence and the main sequence share a continuous PE formulation, two issues arise. First, the continuous temporal indices force the model to interpret the reference images as immediate historical frames. Consequently, the first generated frame tends to exhibit transition artifacts as it attempts to interpolate from the reference views. Second, the shared contiguous PE scale degrades the model's capability to distinguish between the conditioning reference priors and the target generation sequence. 
To address these issues, and inspire by Ominicontrol~\cite{ominicontrol,storymem}, we modify the 3D Rotary Positional Embedding (RoPE) by assigning negative, discrete temporal indices to the reference latents. Given $\mathcal{N}$ reference views and $\mathcal{M}$ target frames, the temporal positional indices are formulated as: 
\begin{equation}
    \{-N \cdot g, -(N-1) \cdot g, \dots, -g, 0, 1, \dots, M-1\}
\end{equation}
where $g$ is a predefined temporal interval gap. By utilizing negative indices with an explicit gap, we preserve the zero-based temporal encoding for the target video $x$. Concurrently, the reference views are explicitly separated and embedded as global spatial conditions, preventing the network from treating them as adjacent historical frames. 

\subsection{Dataset Construction}
\label{sec:dataset}

Training \methodname requires a dataset that provides multi-view reference images of objects, high-quality target video sequences with camera motions conforming to real-world distributions, and accurate 3D pose annotations. However, existing datasets exhibit structural limitations that hinder their direct application to this task. 

Current general video datasets, such as OpenVid~\cite{openvid} and Koala36M~\cite{koala36m}, provide dynamic scenes but lack 3D proxy annotations and predominantly feature stationary cameras. Meanwhile, existing real-world object-centric datasets (\eg, CO3D~\cite{co3d}, OmniObject3D~\cite{omniobject3d}, and MVImageNet~\cite{mvimagenet}) provide multi-view observations but typically operate under static lighting conditions with simple orbital camera motions, diverging from the varied trajectories observed in standard videos. Furthermore, 3D asset repositories such as Objectverse~\cite{objectverse} and Texverse~\cite{texverse} contain detailed geometries and textures, but they consist of uncurated static models that necessitate extensive filtering and rendering pipelines to formulate the paired video sequences. Recently, multi-camera synthetic datasets like SynCamVideo~\cite{syncammaster} and OmniWorld~\cite{omniworld} have been proposed to offer multi-view content with annotated camera trajectories; however, their primary focus lies on human subjects or large-scale scenes rather than object-centric scenarios.

To address this data scarcity, we construct a mixed dataset comprising data from multiple domains, including synthetic rendering from 3D assets, real-world extraction and high-fidelity AI-generated. This multi-source strategy mitigates the biases inherent in single-domain data, ensuring a balanced distribution of object categories, variable environmental lighting, and diverse camera trajectories.

\textbf{Synthetic Rendering.} 
We construct a synthetic dataset by stochastically composing diverse assets via Blender. Specifically, the 3D object models are sourced from Google Scanned Objects~\cite{downs2022googlescannedobjectshighquality} and TexVerse~\cite{texverse}, while the environmental backgrounds are sampled from a curated collection of over 40 scene templates and 100 high-dynamic-range imaging (HDRI) maps. This rendering pipeline explicitly outputs object videos $x$ featuring random camera motions, alongside the perfectly aligned multi-view reference images and their precise spatial poses.

\textbf{Real-World Extraction and AI-Generated.} 
To capture complex real-world dynamics, we extract an object-centric subset from the OpenVidHD~\cite{openvid} and CO3D~\cite{co3d} dataset. For each target video, we employ SAM3~\cite{sam3} to segment the object of interest and utilize ReconViaGen to reconstruct its corresponding 3D proxy. Subsequently, FoundationPose is applied to estimate the frame-wise spatial configurations. To construct the multi-view reference images, we heuristically select a subset of video frames that provide the most comprehensive viewpoint coverage of the target object. Furthermore, we also utilize Wan2.2 to synthesize high-quality, highly diverse object-centric videos. These generated videos are then processed through the identical extraction and annotation pipeline as the real-world data (i.e., SAM3~\cite{sam3}, ReconViaGen~\cite{reconviagen}, and FoundationPose~\cite{wen2024foundationposeunified6dpose}) to obtain the reference images, corresponding 3D proxy, and object poses.

\begin{figure}[tb!]
    \centering
    \includegraphics[width=0.999\linewidth]{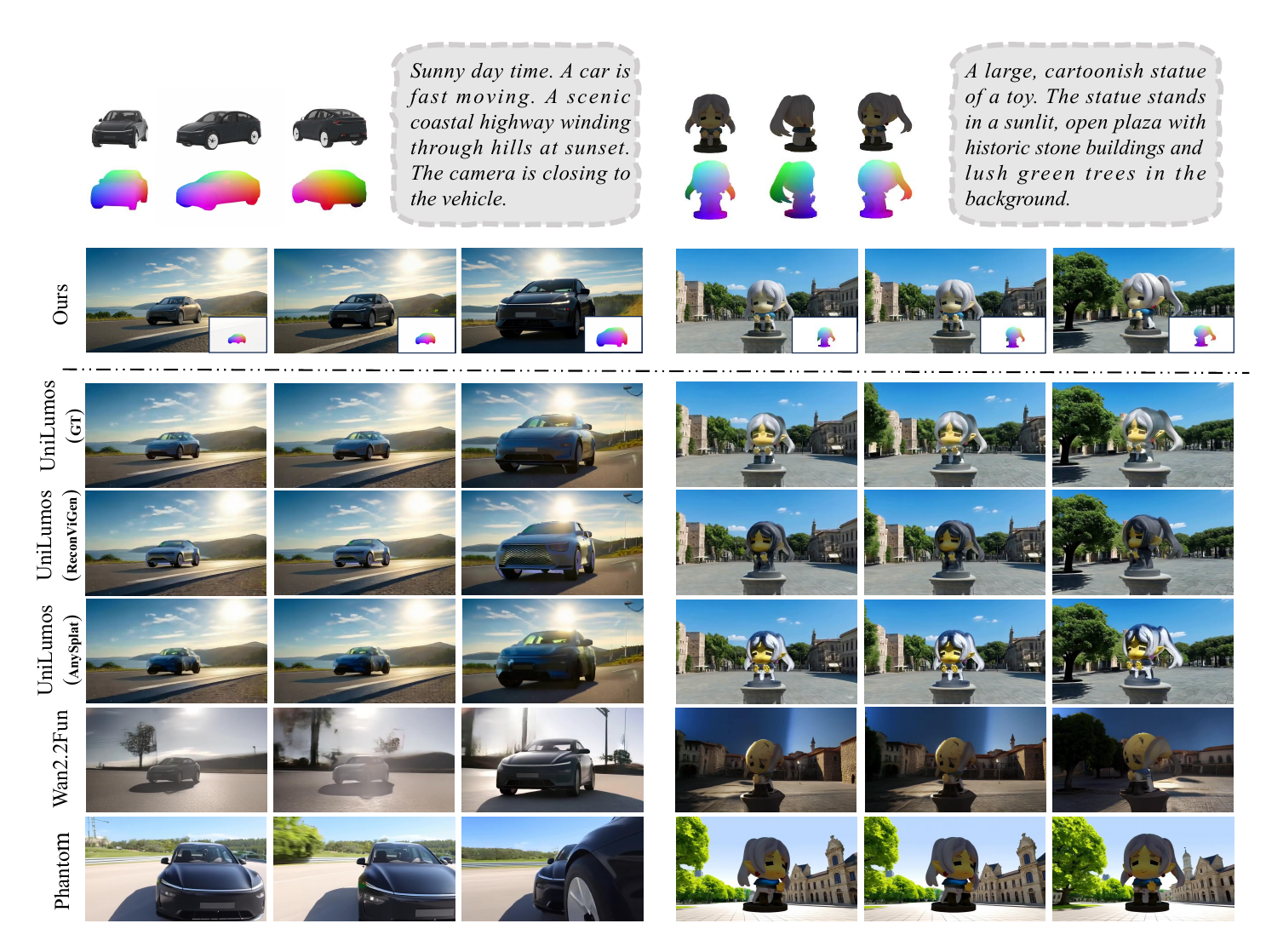}
    \caption{Comparision of rendering quality and lighting consistency.}
    \label{fig:exp1}
    \vspace{-5mm}
\end{figure}

\section{Experiments}
\label{sec:experiments}

The experimental evaluation validates \methodname on diverse synthetic and real-world benchmarks. We assess the framework across several key dimensions: Rendering Quality and Lighting Consistency (Sec.~\ref{sec:exp_quality}), and the Multi-view Consistency of rendered objects (Sec.~\ref{sec:exp_multiview}). Furthermore, we perform ablation studies to verify the efficacy of specific architectural components (Sec.~\ref{sec:exp_ablation}) and demonstrate practical downstream applications (Sec.~\ref{sec:exp_app}). Extensive additional experiment and ablations are deferred to the supplementary material.


\subsection{Implementation Details}
\label{sec:implementation_details}

We implement \methodname based on the pre-trained \textit{Wan2.2Fun 5B Ref Control}~\cite{videoxfun} architecture. The training process is conducted on 16 NVIDIA A800 GPUs for a total of 20,000 steps with a batch size of 32, utilizing the AdamW optimizer. Specifically, the model is trained on video sequences featuring a spatial resolution of $480 \times 832$ and a temporal length of 81 frames. During the training phase, to construct the multi-view reference $\mathcal{I}$, we randomly sample 3 to 8 images from the reference sources. To further enhance the model's robustness and prevent overfitting to specific spatial arrangements, we apply several targeted augmentation strategies: we randomly shuffle the sequence order of the reference views and introduce slight random scaling as well as spatial translation to the positional coordinates of the 3D proxy.

\subsection{Rendering Quality and Lighting Consistency}
\label{sec:exp_quality}

This section evaluates the proposed method against state-of-the-art approaches in terms of rendering quality and lighting consistency under novel viewpoints and environments. Given the absence of identical end-to-end frameworks for this specific task, we construct representative proxy baselines categorized into mainstream two-stage ``reconstruct-then-render'' pipelines and subject-driven video generation models.

\textbf{Baselines.} We select the baselines based on their relevance to object rendering and spatial composition, categorizing them into mainstream two-stage pipelines and subject-driven generative models. For the two-stage pipelines, which explicitly separate 3D modeling from environmental relighting, we select UniLumos~\cite{unilumos} as the standard compositing and relighting backbone. To decouple reconstruction errors from relighting performance, we construct three variants: (1) \textbf{UniLumos(GT)}, which utilizes ground-truth geometry and foreground textures to establish the upper bound of the relighting module; (2) \textbf{UniLumos(ReconViGen~\cite{reconviagen})}, coupled with generative 3D priors; and (3) \textbf{UniLumos(AnySplat~\cite{anysplat})}, integrating feed-forward 3DGS for explicit reconstruction. To represent the alternative paradigm of 2D reference-based generation, we employ Phantom~\cite{phantom} and \textit{Wan2.2Fun 5B Ref Control}~\cite{videoxfun}. These subject-driven models utilize visual conditioning to inject object features into new backgrounds. As they lack intrinsic support for multi-view inputs, we select a single optimal reference view from $\mathcal{I}$ as their visual condition.

\textbf{Evaluation.} We employ dataset-specific evaluation metrics. For synthetic data with available ground truth, we compute PSNR and SSIM to evaluate pixel-level fidelity. For real-world captures lacking ground truth, we utilize VBench++~\cite{vbench} to assess perceptual quality, temporal consistency, and illumination naturalness.

\textbf{Results.} As shown in Tab.~\ref{tab:test_merged} and Fig.~\ref{fig:exp1}, \methodname consistently outperforms all baselines across both physical-structural and perceptual metrics. 
Visually, two-stage pipelines exhibit distinct limitations in complex relighting: UniLumos(GT) presents observable color shifts despite using ground-truth foregrounds, while variants relying on practical 3D reconstruction suffer from further degradation due to error accumulation. 
For the subject-driven generation paradigm, Phantom leverages the priors of video diffusion models to yield competitive smoothness and aesthetic scores. However, lacking explicit 3D geometric constraints, it cannot maintain strict multi-view consistency or adhere to precise camera trajectories. This spatial deviation directly contributes to its lower PSNR and SSIM scores. By integrating explicit 3D proxy guidance with 2D visual priors, \methodname avoids the structural deviations of 2D models and the lighting artifacts of the two-stage pipelines, ensuring consistent high-fidelity rendering.

\begin{table}[t]
  \caption{Quantitative comparison of Rendering Quality and Lighting Consistency. \textbf{Bold} number indicate the best performance.}
  \label{tab:test_merged}
  \centering
  \small
  \setlength{\tabcolsep}{3pt}
  \renewcommand{\arraystretch}{1.15}
  \begin{tabular}{ccccccc}
    \toprule
    Method & PSNR$\uparrow$ & SSIM$\uparrow$ & 
     \begin{tabular}[c]{@{}c@{}}Text\\Align$\uparrow$\end{tabular}&
    \begin{tabular}[c]{@{}c@{}}Motion\\Smoothness$\uparrow$\end{tabular} &
    \begin{tabular}[c]{@{}c@{}}Aesthetic\\Quality$\uparrow$\end{tabular} &
    \begin{tabular}[c]{@{}c@{}}Imaging\\Quality$\uparrow$\end{tabular} \\
    \midrule
    Phantom~\cite{phantom}        & 11.400 & 0.514 &24.620& 0.984 & 0.568 & 0.709 \\
    UniLumos(GT)~\cite{unilumos}  & 12.060 & 0.585 &25.630& 0.978 & 0.569 & 0.669 \\
    UniLumos(ReconViGen~\cite{reconviagen}) & 11.010& 0.338 &25.870& 0.975 & 0.571 & 0.666 \\
    UniLumos(AnySplat~\cite{anysplat}) & 11.160& 0.362 &25.930& 0.975 & 0.557 & 0.624 \\
    Wan2.2Fun      & 14.300 & 0.641 &25.250& 0.984 & 0.530 & 0.601 \\
    Ours           & \textbf{18.260} & \textbf{0.684} & \textbf{27.720}& \textbf{0.988} & \textbf{0.573} & \textbf{0.712} \\
    \bottomrule
  \end{tabular}
  \vspace{-4mm}
\end{table}

\subsection{Multi-view Consistency}
\label{sec:exp_multiview}

\begin{table}[t]
  \caption{Quantitative comparison of Multi-view Consistency and Reconstruction. \textbf{Bold} numbers indicate the best performance.}
  \label{tab:test_clip_dino}
  \centering
  \small
  \setlength{\tabcolsep}{6pt} 
  \renewcommand{\arraystretch}{1.2}
  \begin{tabular}{l cc | cc}
    \toprule
    Method & \multicolumn{2}{c|}{CLIP} & \multicolumn{2}{c}{DINO} \\
    \cmidrule(lr){2-3} \cmidrule(lr){4-5}
    & Img/Avg.$\uparrow$ & Img/Max.$\uparrow$ & Avg.$\uparrow$ & Max.$\uparrow$ \\
    \midrule
    AnySplat    & 0.861 & 0.934 & 0.191 & 0.323 \\
    ReconViaGen & 0.796 & 0.927 & 0.448 & 0.727 \\
    Ours        & \textbf{0.888} & \textbf{0.956} & \textbf{0.725} & \textbf{0.860} \\
    \bottomrule
  \end{tabular}
\end{table}

\begin{figure}[tb!]
    \centering
    \includegraphics[width=0.999\linewidth]{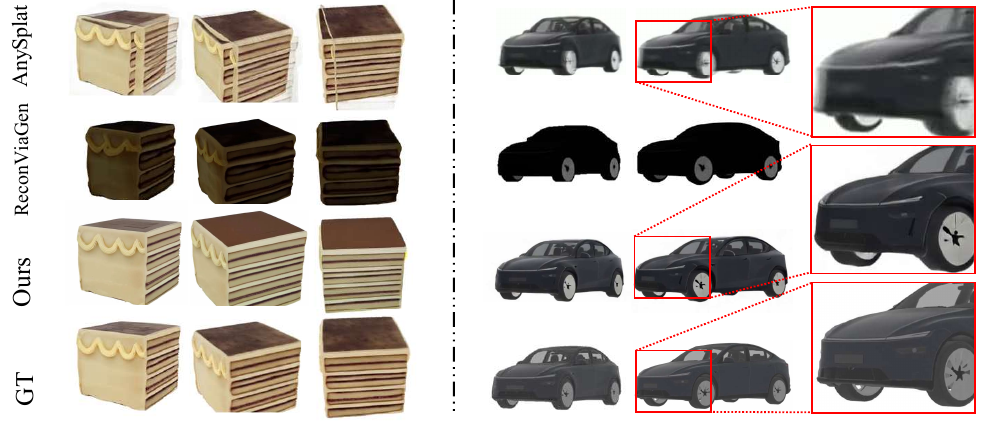}
    \caption{Qualitative comparison of Multi-view Consistency.}
    \label{fig:exp2}
    \vspace{-4mm}
\end{figure}

To evaluate multi-view consistency while decoupling it from the confounding factors of environmental relighting, we isolate the object rendering process against canonical blank backgrounds. This controlled setup strictly assesses the preservation of intrinsic object appearance across varying viewpoints.

\textbf{Baselines.} We construct the baselines to represent two primary paradigms of 3D reconstruction suitable for novel view synthesis. Specifically, we select AnySplat~\cite{anysplat} as the representative of feed-forward explicit 3DGS, and ReconViaGen~\cite{reconviagen} to represent generative 3D modeling. For both baselines, the multi-view reference images are utilized to reconstruct the underlying 3D representations. The reconstructed models are subsequently rendered along the target pose sequence to synthesize the video frames. This setup establishes a direct comparison between conventional 3D reconstruction-based pipelines and our generative rendering framework.

\textbf{Evaluation.} We employ a comprehensive set of metrics to quantify both the perceptual and physical fidelity of the synthesized novel views. To evaluate semantic and structural consistency, we calculate the CLIP~\cite{clip} similarity to measure the high-level semantic alignment between the rendered frames and the multi-view references, while utilizing the DINO~\cite{dino} score to assess fine-grained structural preservation and spatial layout correspondences. Furthermore, we introduce PSNR and SSIM to evaluate pixel-level reconstruction accuracy and geometric fidelity under canonical setups. Detailed calculation procedures and formulations for all utilized metrics are provided in the supplementary material.

\textbf{Results.} As shown in Tab.~\ref{tab:test_clip_dino}. \methodname achieves the best performance across all evaluated metrics. AnySplat exhibits a noticeable degradation across the structural (DINO) and physical fidelity metrics (PSNR and SSIM). As shown in Fig.~\ref{fig:exp2}, the 3DGS reconstruction suffers from blurriness and distortion. Conversely, while ReconViaGen demonstrates distinct performance patterns, it falls short in overall semantic alignment and pixel-level accuracy. The synthesized geometry and texture details lack strict consistency with the reference inputs. By integrating explicit 3D proxy guidance with 2D diffusion priors, our method circumvents the blurriness of explicit 3DGS reconstruction and the geometric-texture inconsistencies of generative pipelines, thereby maintaining robust identity preservation and high-fidelity rendering.

\subsection{Ablation Study}
\label{sec:exp_ablation}

\begin{figure}[t]
  \centering
  \small
  \begin{minipage}{0.55\textwidth}
    \centering
    \includegraphics[width=\textwidth]{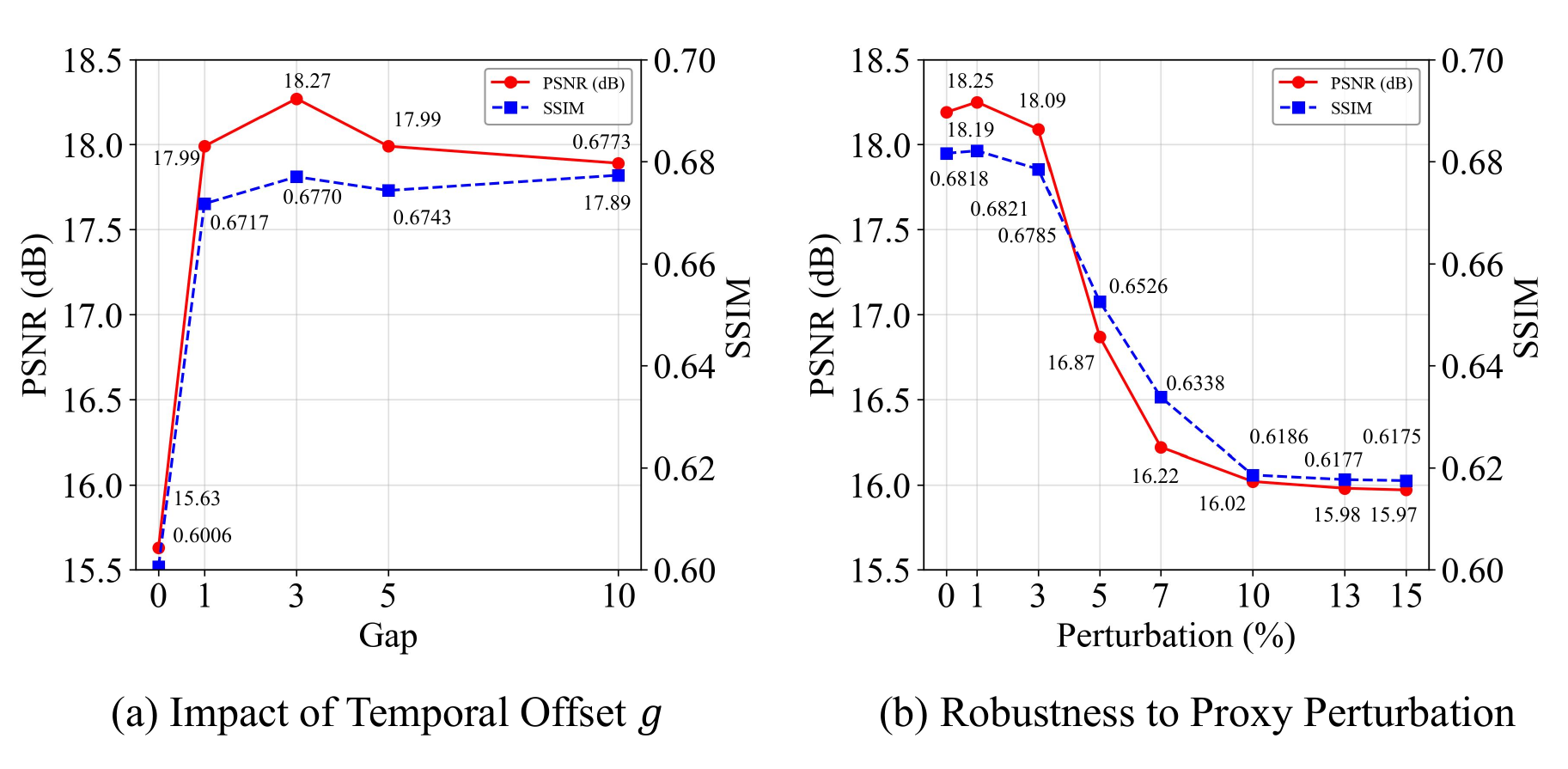}
    \caption{\textbf{Quantitative Ablation Results.} (a) Impact of the temporal offset $g$ on rendering quality. A sufficient gap ($g=3$) effectively isolates spatial priors. (b) Model robustness under varying degrees of spatial perturbations.}
    \label{fig:ablation}
  \end{minipage}
  \hspace{0.05\textwidth}
  \begin{minipage}{0.38\textwidth}
    \centering
    \includegraphics[width=\textwidth]{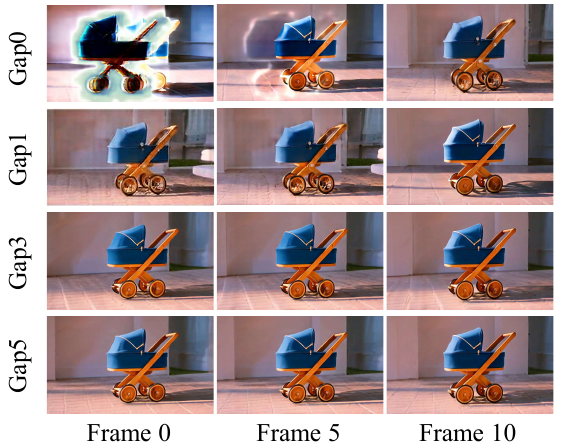}
    \caption{Qualitative ablation results of temporal offset.}
    \label{fig:ablation1}
  \end{minipage}
\end{figure}

\textbf{Impact of Temporal Offset ($g$).} 
To evaluate the sensitivity of the negative temporal indices $g$, we test various values of $g \in \{0, 1, 3, 5, 10\}$. As illustrated in Fig.~\ref{fig:ablation}(a), setting $g=0$ results in a severe performance drop (PSNR drops to 15.63). As shown in Fig~\ref{fig:ablation1} without a sufficient temporal buffer, the pre-trained DiT natively attempts to interpolate between the static reference view and the first target frame, leading to noticeable smearing artifacts. However, simply introducing a gap of $g=1$ significantly recovers the performance. We observe that $g=3$ achieves the optimal quantitative results, effectively isolating the spatial priors while maintaining strong conditional guidance. 

\textbf{Robustness to Proxy Quality.}
To evaluate the robustness of \methodname, we simulate tracking inaccuracies by injecting varying levels of random spatial perturbations into the target coordinate maps during inference. As shown in Fig.~\ref{fig:ablation}(b), the model maintains stable performance under minor spatial noise, demonstrating its capacity to absorb slight geometric jitters. As the perturbation level increases, the evaluation metrics exhibit a graceful degradation rather than a sharp collapse. Under severe noise injection, the performance stabilizes into a distinct plateau. This trend suggests that when the 3D proxy becomes heavily corrupted, the model reduces its reliance on unreliable geometric cues and instead depends on the 2D diffusion prior to synthesize a plausible appearance, thereby avoiding severe rendering failures.

\subsection{Applications}
\label{sec:exp_app}

\begin{figure}[t]
    \centering
    \includegraphics[width=0.999\linewidth]{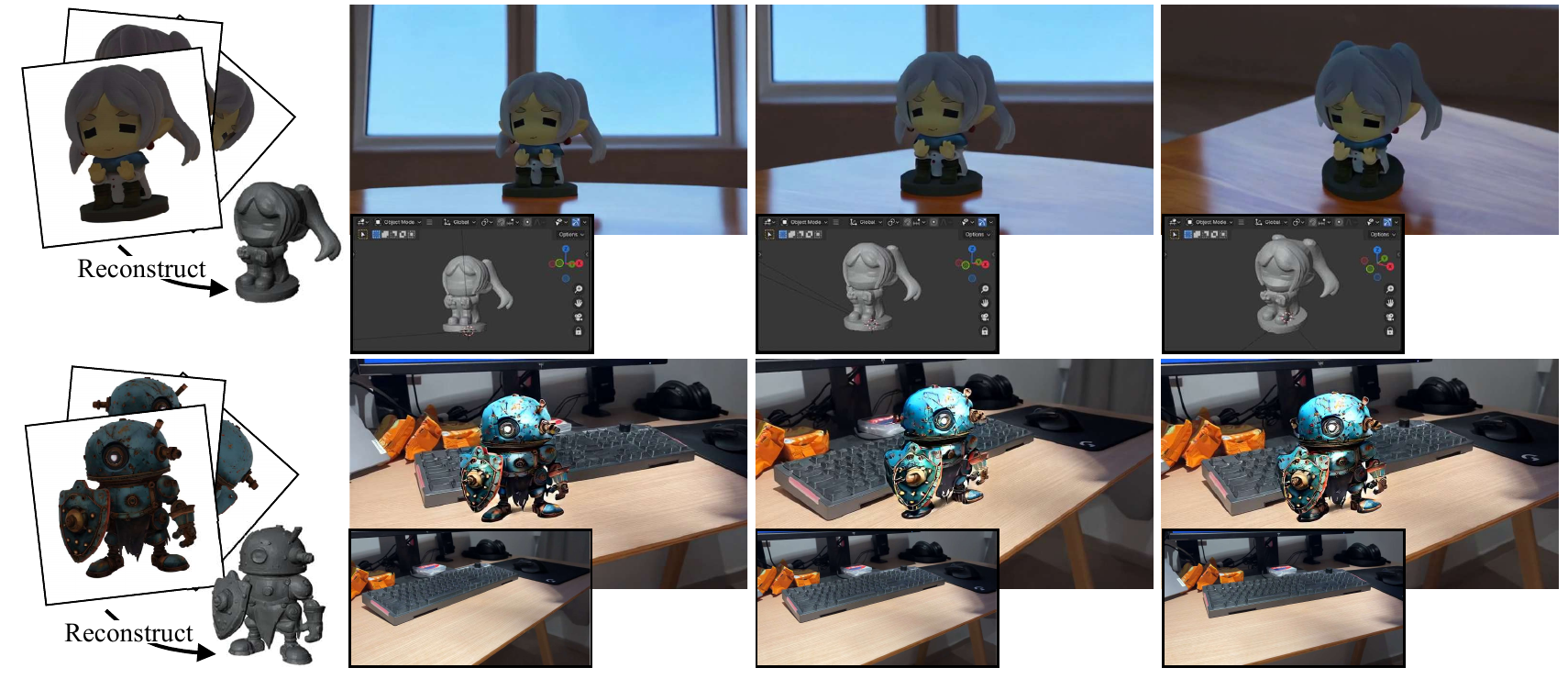}
    \caption{Application of our \methodname.}
    \label{fig:application}
    \vspace{-3mm}
\end{figure}

We demonstrate the practical utility of \methodname in two downstream scenarios: offline rendering and object insertion, as depicted in Fig.~\ref{fig:application}. On the the first row, we deploy \methodname as an offline renderer plug-in within Blender. It synthesizes high-fidelity object animations driven by text prompts $c$ and explicit camera trajectories exported directly from the 3D editor software. On the second row, we showcase the seamless insertion of 3D models into real-world video sequences. The inserted objects naturally blend into the physical scenes, exhibiting highly realistic physical interactions, including accurate reflections and plausible shadows.

\section{Limitation}
\label{sec:limit}

\begin{figure}
    \centering
    \includegraphics[width=0.999\linewidth]{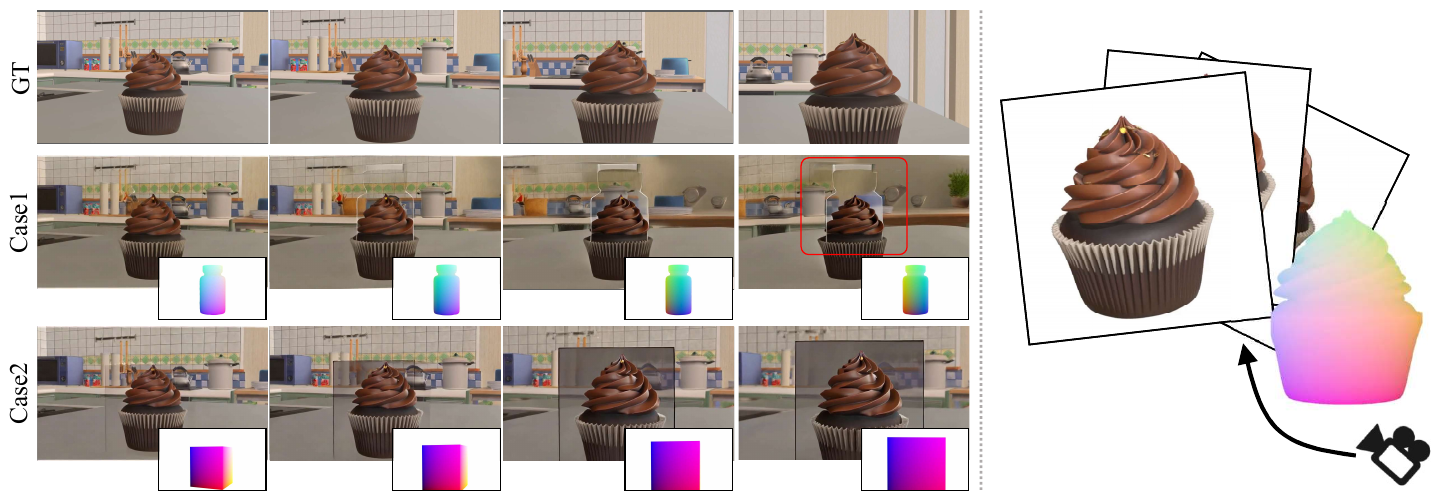}
    \caption{\textbf{Failure cases.} The top row (GT) displays the ground-truth sequence. In Case 1 and Case 2, the model is conditioned on the multi-view appearances of a cupcake, but provided with the explicit spatial coordinate maps of a bottle and a cube, respectively. This geometric inconsistency prevents the model from reconciling the conflicting spatial and appearance conditions, leading to structural artifacts and degraded rendering fidelity.}
    \label{fig:limitations}
    \vspace{-2mm}
\end{figure}

Although our ablation studies demonstrate that \methodname exhibits robustness to spatial perturbations by leveraging 2D diffusion priors, the accuracy and structural fidelity of the underlying 3D proxy still impact the final rendering results. Specifically, as illustrated in Fig.~\ref{fig:limitations}, utilizing an incorrect or structurally mismatched 3D proxy compromises the generative output. Under such conditions, the model struggles to reconcile the conflicting spatial conditions with the reference appearances, leading to visual artifacts and a quantifiable degradation in viewpoint control accuracy. Therefore, acquiring corresponding 3D geometries remains a prerequisite for strictly controlled generation, which we aim to address in future work by integrating more advanced proxy extraction modules.

\section{Conclusion}
\label{sec:conclusion}
We propose \methodname, a unified framework integrating the reconstructed 3D proxies to guide the video generative models to achieve high-quality object rendering on arbitrary viewpoints under arbitrary lighting conditions. 
Our method not only enjoys the accurate viewpoint control using the reconstructed 3D proxy but also enables high-quality rendering in different lighting environments using diffusion generative models without explicitly modeling complex materials and lighting.

\subsection*{Acknowledgments}
We also thank Julian Tanke from Sony and Yanpei Cao from VAST for the valuable guidance and comments throughout the development of this work.


%
%
\bibliographystyle{splncs04}
\bibliography{main}
\end{document}